\documentclass[letterpaper, 10 pt, conference]{ieeeconf} 
\IEEEoverridecommandlockouts                
\usepackage{graphics} 
\usepackage{epsfig} 
\usepackage{times} 
\usepackage{amsmath} 
\usepackage{amssymb}  
\usepackage{hyperref}
\usepackage{upgreek}
\usepackage{cite}
\usepackage{cleveref}
\usepackage{graphicx}
\usepackage{relsize}
\usepackage{longtable}
\usepackage{color,soul}
\usepackage{gensymb}
\usepackage{url}
\usepackage{array}
\usepackage{multicol} 
\usepackage{siunitx}
\usepackage{subcaption}
\usepackage{multirow}
\usepackage{makecell}
\usepackage{ctable}
\usepackage{arydshln}

\usepackage{amsthm}
\newtheorem*{remark}{Remark}

\title{\LARGE \bf
Turning Circle-based Control Barrier Function for Efficient Collision Avoidance of Nonholonomic Vehicles}

\author{Changyu Lee, Kiyong Park, and Jinwhan Kim* 
\thanks{Manuscript received 30 April 2021; accepted 12 September 2021. Date of publication 4 October 2021; date of current version 17 October 2022. This letter was recommended for publication by Associate Editor}
\thanks{This research was supported by. \textit{(Corresponding author: Jinwhan Kim)}}
\thanks{The authors are with the Department of Mechanical Engineering, Korea Advanced Institute of Science and Technology, Daejeon 34141, South Korea (e-mail: leeck@kaist.ac.kr; qkrrldyd777@kaist.ac.kr; jinwhan@kaist.ac.kr)}%
}

\begin{document}
\maketitle
\thispagestyle{empty}
\pagestyle{empty}

\everymath{\displaystyle}
\begin{abstract}
This paper presents a new control barrier function (CBF) designed to improve the efficiency of collision avoidance for nonholonomic vehicles. 
Traditional CBFs typically rely on the shortest Euclidean distance to obstacles, overlooking the limited heading change ability of nonholonomic vehicles. This often leads to abrupt maneuvers and excessive speed reductions, which is not desirable and reduces the efficiency of collision avoidance. Our approach addresses these limitations by incorporating the distance to the turning circle, considering the vehicle's limited maneuverability imposed by its nonholonomic constraints.
The proposed CBF is integrated with model predictive control (MPC) to generate more efficient trajectories compared to existing methods that rely solely on Euclidean distance-based CBFs. 
The effectiveness of the proposed method is validated through numerical simulations on unicycle vehicles and experiments with underactuated surface vehicles.
\end{abstract}


\section{Introduction}
Generating collision-free trajectories is essential for mobile robots and autonomous vehicles, and control barrier functions (CBFs) have gained significant research attention as a crucial tool in achieving this goal. CBFs ensure safety by guaranteeing the forward invariance of a defined safe set \cite{ames2016control,ames2019control}. When combined with model predictive control (MPC), which is based on optimal control theory, constraints on state variables can be applied \cite{borrelli2017predictive}, allowing for the determination of integrated planning and control inputs that ensure safe and reliable collision avoidance.

CBF and MPC have been extensively studied and successfully applied in various contexts. MPC has been widely used to enhance navigation and control performance in different types of vehicles including marine vehicles \cite{kim2023navigable,lee2024safety}, quadrotors \cite{lindqvist2021reactive}, and ground vehicles \cite{ammour2022mpc}.
In these applications, CBF has also demonstrated versatility in improving safety for autonomous vehicles \cite{chen2017obstacle}, bipedal robots \cite{agrawal2017discrete}, adaptive cruise control systems \cite{ames2014control}, and multi-agent systems \cite{jankovic2023multiagent}. 
Despite their achievements, both approaches face their own challenges. MPC may struggle with obstacles beyond its prediction horizon, necessitating a careful balance between horizon length and computational efficiency, while CBF approaches can result in overly conservative operational strategies. 

Recent research has explored the combination of CBF and MPC to address these limitations and leverage the synergy between them \cite{zeng2021safety}.
This strategy integrates the discrete-time CBF as a collision avoidance constraint within the MPC algorithm, resulting in more optimized solutions and improving obstacle avoidance capabilities, even with short prediction horizons. 
This combined approach has been successfully applied in various scenarios, such as ensuring the safe navigation of social robots around moving humans \cite{vulcano2022safe} and developing collision avoidance strategies for multiple quadrotors \cite{jin2022collision}.
It has also been employed to enable mobile robots to avoid both static and dynamic obstacles by applying the dynamic CBF concept \cite{jian2023dynamic}.

A prevalent strategy in CBF-based obstacle avoidance is to maintain a predetermined safe distance from obstacles, which is defined by the equation $h = d-r$ \cite{vulcano2022safe,jin2022collision,manjunath2021safe,peng2023safety,bruggemann2022simultaneous, jian2023dynamic}. Where $d$ is the Euclidean distance to the obstacle and $r$ represents the safety margin. This margin accounts for both the vehicle's size and the obstacle's shape. 
The primary goal of the Euclidean distance-based CBF (ED-CBF) method is to ensure that the value of $h$ remains non-negative, thereby maintaining a safe distance from obstacles. This is accomplished through the condition $\dot{h} + \alpha (h) \geq 0$, where $\alpha$ is an extended class $\mathcal{K}$ function\footnote{A continuous function $\alpha:(-b, a) \rightarrow (-\infty, \infty)$, where $a, b>0$, is classified as belonging to the extended class $\mathcal{K}$ if it is strictly increasing and satisfies $\alpha(0)=0$.} \cite{ames2016control}. 
While this approach is effective for holonomic vehicles, it fails to account for relative heading and nonholonomic motion constraints. 
This oversight can result in unnecessary or overly aggressive avoidance maneuvers, compromising the efficiency of collision avoidance.
Therefore, it is essential to develop improvements in the CBF design that account for the nonholonomic characteristics of vehicles.

To overcome the limitations of ED-CBF methods, our study proposes a turning circle-based CBF (TC-CBF) to enhance collision avoidance efficiency in nonholonomic vehicles. This approach develops a CBF based on the proximity of the vehicle's turning circles to surrounding obstacles. These turning circles are calculated by considering the vehicle's current heading, speed, and maximum turn rates, ensuring feasible collision avoidance in either direction while accounting for the vehicle's nonholonomic characteristics.
The TC-CBF is then integrated with MPC to generate efficient collision avoidance trajectories. 
This integrated approach is benchmarked against an existing MPC framework employing the ED-CBF method through both numerical simulations and real-world validations. Obstacle avoidance experiments were conducted in static and dynamic environments using unicycles and autonomous surface vehicles (ASVs). 
The main contributions of this study can be summarized as follows:
\begin{itemize}
\item A new CBF is proposed to improve the collision avoidance efficiency of nonholonomic vehicles. The proposed CBF is based on the proximity of the vehicle's turning circles to nearby objects, effectively integrating the vehicle's heading and turning capabilities into the CBF formulation. 
\item The MPC-TCCBF framework is proposed to generate efficient trajectories by utilizing MPC with TC-CBF as obstacle avoidance constraints.
\item Experimental results demonstrate that the proposed framework outperforms the existing ED-CBF-based MPC in navigating both static and dynamic environments for nonholonomic vehicles, especially unicycles and ASVs.
\end{itemize}

The following section outlines the preliminaries. Section~\ref{section3} provides a detailed formulation of our proposed approach, including the CBF design and trajectory planning algorithm. Section~\ref{section4} presents the simulation and experimental results. Finally, the conclusions of this study are summarized in Section~\ref{section5}.

\section{Preliminaries} \label{sec:section2}
\subsection{Control Barrier Functions} \label{sec:subsec}
Consider a nonlinear control-affine system:
\begin{equation} \label{eq:system}
    \dot{\mathbf{x}} = f(\mathbf{x}) + g(\mathbf{x})\mathbf{u},
\end{equation}
where $\mathbf{x}\in \mathcal{X}\subset \mathbb{R}^n$ and $\mathbf{u}\in \mathcal{U} \subset \mathbb{R}^m$ represent the state and control input vectors, respectively.
The set $\mathcal{S} \subset \mathbb{R}^n$ defines the safe region within which the vehicle's states must remain. The objective is to design a controller that guarantees the forward invariance of the set $\mathcal{S}$. This ensures that if $\mathbf{x}(0) \in \mathcal{S}$, then $\mathbf{x}(t) \in \mathcal{S}$ for all $t \geq 0$, ensuring the vehicle consistently remains in a safe state. Forward invariance of the safe set can be achieved through the use of a CBF \cite{ames2016control}.

Let $\mathcal{S}$ be defined as the superlevel set of a continuously differentiable function $h: \mathbb{R}^n \rightarrow \mathbb{R}$, formulated as:
\begin{equation}
\begin{aligned}
\mathcal{S} & =\left\{\mathbf{x} \in \mathbb{R}^n: h(\mathbf{x}) \geq 0\right\}, \\
\partial \mathcal{S} & =\left\{\mathbf{x} \in \mathbb{R}^n: h(\mathbf{x})=0\right\}, \\
\operatorname{Int}(\mathcal{S}) & =\left\{\mathbf{x} \in \mathbb{R}^n: h(\mathbf{x})>0\right\} .
\end{aligned}
\end{equation}
The function $h$ is a CBF if $ \textstyle \frac{\partial h}{\partial \mathbf{x}}\mathbf{x} \neq 0 $ for all $\mathbf{x} \in \partial \mathcal{S}$, and there exists an extended class $\mathcal{K}$ function $\alpha$ such that for the system \eqref{eq:system} and for all $\mathbf{x} \in \mathcal{S}$ \cite{ames2019control}:
\begin{equation} \label{eq:cbfconstraints}
\sup _{\mathbf{u} \in \mathcal{U}}[L_f h(\mathbf{x})+L_g h(\mathbf{x}) \mathbf{u}+{\alpha}(h(\mathbf{x}))] \geq 0, 
\end{equation}
where $\textstyle L_f h(\mathbf{x})=\frac{\partial h(\mathbf{x})}{\partial \mathbf{x}} f(\mathbf{x})$ and $\textstyle L_g h(\mathbf{x})=\frac{\partial h(\mathbf{x})}{\partial \mathbf{x}} g(\mathbf{x})$ denote the Lie derivatives of $h(\mathbf{x})$ along $f$ and $g$, respectively.
The admissible control space $U(\mathbf{x})$ is defined as:
\begin{equation}
U(\mathbf{x})=\left\{\mathbf{u} \in \mathcal{U} : L_f h(\mathbf{x})+L_g h(\mathbf{x}) \mathbf{u}+\alpha(h(\mathbf{x})) \geq 0\right\},
\end{equation}
ensuring the forward invariance of $\mathcal{S}$.

The choice of the extended class $\mathcal{K}$ function $\alpha$ influences how the state approaches the boundary of $\mathcal{S}$. 
A common choice for $\alpha(h)$ is a scalar multiple of $h(\mathbf{x})$, typically $\alpha(h(\mathbf{x}))= \gamma h(\mathbf{x})$ with $\gamma>0$.
This condition requires the control input to satisfy:
\begin{equation} 
\label{eq:cbf_ic}
L_f h(\mathbf{x})+L_g h(\mathbf{x}) \mathbf{u}+\gamma h(\mathbf{x}) \geq 0.
\end{equation}

This condition can be extended to the discrete-time domain as follows \cite{zeng2021safety}:
\begin{equation}\label{eq:dcbf_intro}
    \Delta {h}(\mathbf{x}_k,\mathbf{u}_k) \geq -\gamma h(\mathbf{x}_k), \ 0< \gamma  \leq 1,
\end{equation}
where $\Delta {h}(\mathbf{x}_k,\mathbf{u}_k) = {h}(\mathbf{x}_{k+1}) - {h}(\mathbf{x}_k)$. By satisfying the constraint in \eqref{eq:dcbf_intro}, we ensure that ${h}(\mathbf{x}_{k+1}) \geq (1-\gamma) {h}(\mathbf{x}_{k})$, indicating that the lower bound of the CBF decreases at an exponential rate of $(1-\gamma)$. The discrete-time CBF can be effectively integrated with MPC (MPC-CBF) as an alternative to traditional distance-based collision avoidance constraints, which typically mandate maintaining a minimum distance above a specified safety threshold. This integration has demonstrated comparable performance even with a short prediction horizon \cite{zeng2021safety}.

\subsection{Collision Avoidance using ED-CBF}
\begin{figure}[t]   \centerline{\includegraphics[width=0.75\linewidth]{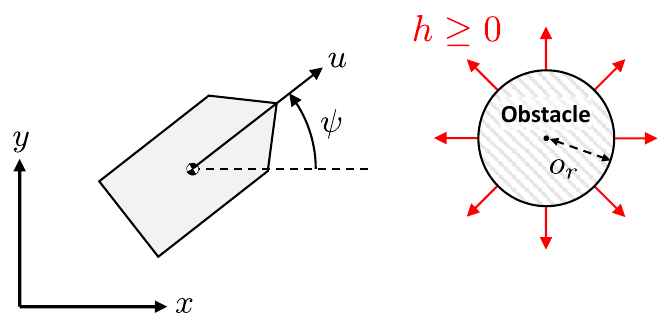}}
    \caption[Coordinate.]{The coordinate systems.}
    \label{fig:coord}
\end{figure}

Consider the nonholonomic unicycle model defined as follows:
\begin{equation}  
    \dot{x} = u\cos\psi, \ 
    \dot{y} = u\sin\psi, \
    \dot{\psi} = r, \
    \dot{u} = a,  
    \label{unicycle}
\end{equation}
where $\mathbf{u} = [r,a]^\top \in \mathcal{U}$ represents the control inputs, with $r$ being the turning rate and $a$ the forward acceleration. $x$, $y$, and $\psi$ denote the vehicle's position and heading, respectively, as illustrated in Fig.~\ref{fig:coord}.

The CBF for obstacle avoidance is formulated as:
\begin{equation} \label{eq:distcbf}
    h(\mathbf{x}) = \sqrt{(x-o_x)^2 + (y-o_y)^2} - (o_r+R_s),
\end{equation}
where $o_x$ and $o_y$ represent the obstacle's position, $o_r$ is the obstacle's radius, and $R_s$ is the safety radius of the vehicle. The positivity of the CBF ensures that the vehicle avoids collisions with obstacles. However, while \eqref{eq:distcbf} is a position-based function that incorporates velocity in its derivation, the vehicle's control inputs are acceleration and turning rate.
To bridge this gap, a higher-order CBF is introduced as follows \cite{nguyen2016exponential}:
\begin{equation}
    \label{eq:edcbf}
    h_e(\mathbf{x}) = \dot{h}(\mathbf{x}) + \alpha h(\mathbf{x}) \geq 0, \ \alpha > 0,
\end{equation}
where
\begin{equation}\label{eq:hod}
    \displaystyle
    \dot{h}(\mathbf{x}) = \frac{(x-o_x)\dot{x} + (y-o_y)\dot{y}}{\sqrt{(x-o_x)^2 + (y-o_y)^2}}.
\end{equation}
In this paper, we will refer to $h_e$ as the ED-CBF.
The ED-CBF can be integrated with MPC (MPC-EDCBF) to replace distance-based obstacle avoidance constraints, such as $h(\mathbf{x})\geq 0$. The obstacle avoidance constraints using ED-CBF can be formulated as follows (for a detailed formulation of MPC-CBF, see \cite{zeng2021safety}):
\begin{equation}
    \label{eq:cbfcon}
    \Delta {h}_e(\mathbf{x}_k,\mathbf{u}_k) \geq -\alpha_e h_e(\mathbf{x}_k), \ 0< \alpha_e  \leq 1.
\end{equation}

\begin{figure}[t]
    \centerline{\includegraphics[width=0.7\linewidth]{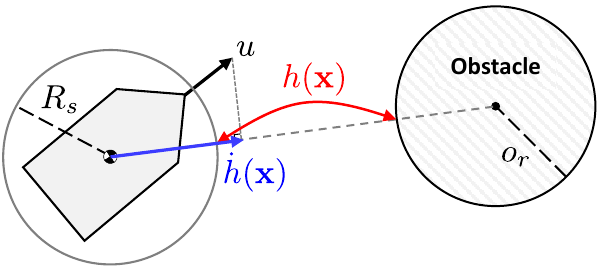}}
    \caption[]{Geometrical representation of the ED-CBF constraints.}
    \label{fig:edcbf}
\end{figure}

The ED-CBF constraints imply that the maximum velocity is limited based on the relative distance to an obstacle, as shown in Fig.~\ref{fig:edcbf}.
However, this approach does not fully account for the vehicle’s turning maneuverability.
This limitation arises because the constraint is applied solely to the projected velocity, which can lead to unnecessary deceleration and aggressive avoidance maneuvers. 

\section{Turning Circle-based Control Barrier Function} \label{section3}
\subsection{Control Barrier Function Design}
\begin{figure}[t]
    \centerline{\includegraphics[width=0.8\linewidth]{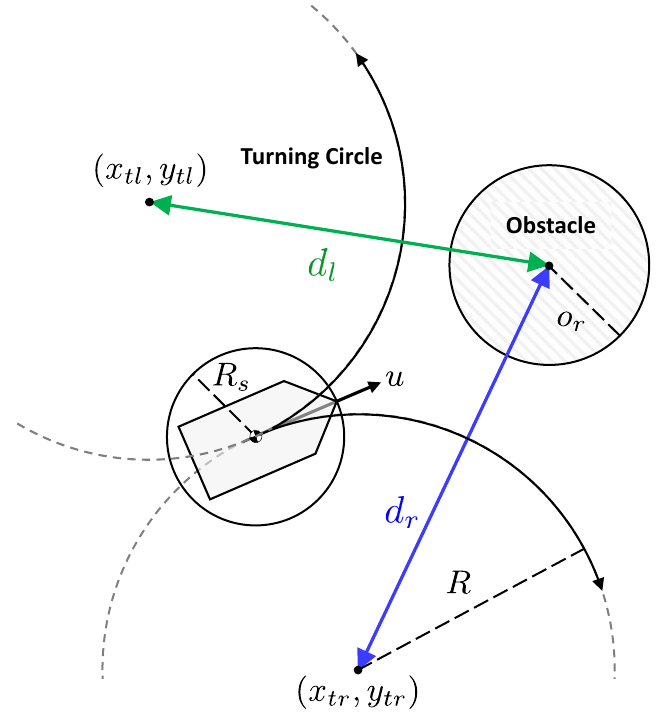}}
    \caption[]{Geometrical representation of the TC-CBF constraints.}
    \label{fig:tccbf}
\end{figure}

The TC-CBF is proposed to address the limitations of ED-CBF in collision avoidance. The TC-CBF is based on the principle that a vehicle can avoid obstacles as long as its turning circle, determined by its maximum turning capability, current speed, and heading, does not intersect obstacles.
The TC-CBF $h_t(\mathbf{x})$ is defined as:
\begin{equation}
    h_t(\mathbf{x}) = \max(h_{tr}(\mathbf{x}), h_{tl}(\mathbf{x})),
    \label{eq:tccbf2}
\end{equation}
where $h_{tr}$ and $h_{tl}$ represent the CBFs for the right and left turning circles, respectively, defined as follows:
\begin{equation}
\begin{aligned}
    h_{tr} (\mathbf{x}) &= d_r - (o_r+R_s+ R), \\
    h_{tl} (\mathbf{x}) &= d_l - (o_r+R_s+ R),
    \label{eq:leftright}
\end{aligned}
\end{equation}
where $d_r$ and $d_l$ represent the distances from the centers of the right and left turning circles to the obstacles, respectively, and are given by $\textstyle d_r = \sqrt{(x_{tr}-o_x)^2 + (y_{tr}-o_y)^2}$ and $\textstyle d_l = \sqrt{(x_{tl}-o_x)^2 + (y_{tl}-o_y)^2}$.
Here, $R$ is the radius of the turning circle, defined as:
\begin{equation}
    R = \frac{u}{r_{\text{max}}},
\end{equation}
where $u$ is the speed of the vehicle and $r_{\text{max}} = \max(r)$ denotes the maximum turning rate. 
The center positions of the turning circles, $(x_{tr},y_{tr})$ for the right and $(x_{tl},y_{tl})$ for the left, are given by:
\begin{equation}
\begin{aligned}    
    (x_{tr},y_{tr})  &= (x + R\cos(\psi-\pi/2), y+R\sin(\psi-\pi/2)), \\
    (x_{tl},y_{tl})  &= (x + R\cos(\psi+\pi/2), y+R\sin(\psi+\pi/2)).
\end{aligned}
\label{eq:tccbf_form}
\end{equation}
The concept of the TC-CBF is illustrated in Fig.~\ref{fig:tccbf}.
Non-negative values of $h_{tr}$ and $h_{tl}$ indicate that the vehicle's right and left turning circles maintain a safe distance from the obstacles.

The vehicle can ensure safety from obstacles as long as the turning circle in at least one direction does not intersect with the obstacle. In other words, this condition is met if either $h_{tr}$ or $h_{tl}$ remains non-negative.
The formulation in \eqref{eq:tccbf2} is designed to fulfill this requirement. 
To integrate the maximum operation in \eqref{eq:tccbf2} with a numerical optimization problem, the following smooth approximation is introduced \cite{molnar2023composing}:
\begin{equation}
    h_t(\mathbf{x}) = \frac{1}{k} \ln \Big( \big(e^{k h_{tr}(\mathbf{x})} + e^{k h_{tl}(\mathbf{x})} \big)/2 \Big).
    \label{eq:tccbf_smooth}
\end{equation}
where $k>0$ is the smoothing parameter.
If \eqref{eq:tccbf_smooth} yields a non-negative value, then the condition $e^{kh_{tr}(\mathbf{x})} + e^{kh_{tl}(\mathbf{x})}\geq 2$ is satisfied.
For instance, if $h_t(\mathbf{x}) \geq 0 $ and $h_{tr}(\mathbf{x}) < 0 $, then since $0< e^{kh_{tr}(\mathbf{x})} < 1$, it follows that $e^{kh_{tl}(\mathbf{x})} \geq 1$, implying that $h_{tl}(\mathbf{x})$ must be positive.
This logic also applies if $h_t(\mathbf{x})\geq 0$ and $h_{tl}(\mathbf{x}) < 0 $. Thus, maintaining a non-negative TC-CBF ensures that both $h_{tl}$ and $h_{tr}$ cannot be negative simultaneously.

\begin{figure}[t]
    \centerline{\includegraphics[width=\linewidth]{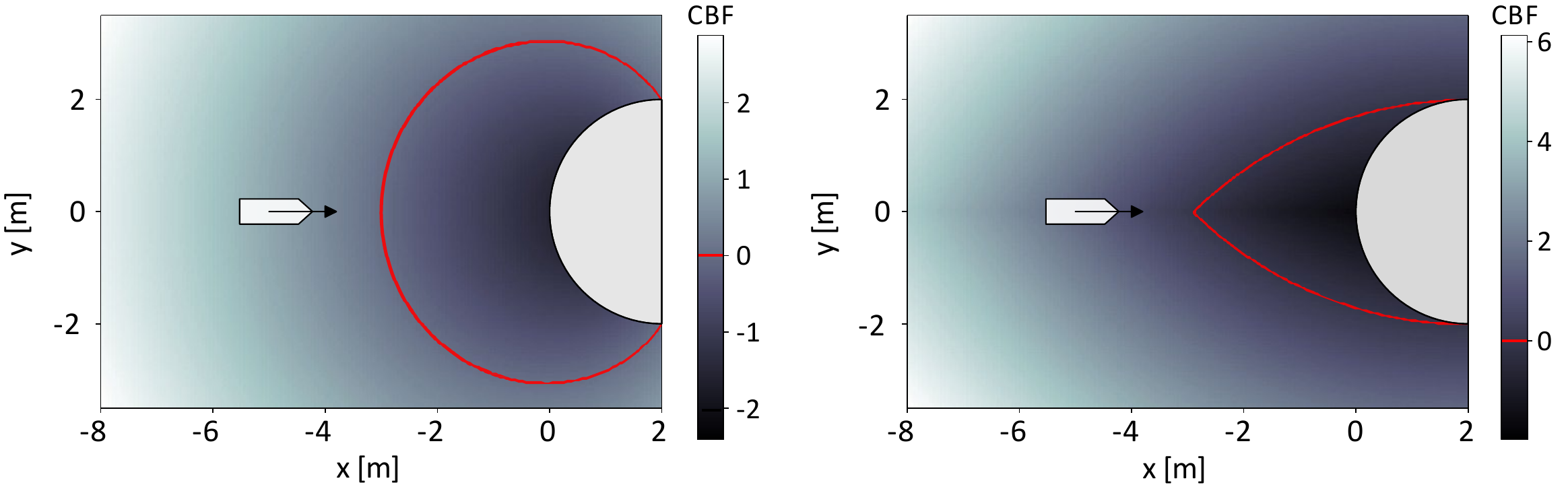}}
    \caption[]{Comparison of the two CBF functions with parameters $r_{\text{max}} = 0.3$, $\alpha = 0.5$, and $u = \SI{1.5}{\meter/s}$.}
    \label{fig:cbfcompare}
\end{figure}

Fig.~\ref{fig:cbfcompare} demonstrates how the ED-CBF and TC-CBF enforce constraints for a circular obstacle at a vehicle speed of $\SI{1.5}{\meter/s}$ and a heading angle of $0^\circ$ across all $x,y$ positions.
Obstacles are depicted in gray, while the red boundary indicates regions where $h_e$ and $h_t$ are zero, meaning that the area within the red boundary must be avoided under the constraints of ED-CBF and TC-CBF to maintain the current speed and heading. 
The size and shape of these restricted regions vary with the vehicle's speed and its orientation relative to the obstacle.
The ED-CBF constraint leads to a wider exclusion zone in the direction perpendicular to the vehicle's heading, necessitating greater rotation or deceleration when avoidance cannot be achieved solely through turning. As a result, vehicles with limited maneuverability might struggle to navigate within the ED-CBF's constraints. Conversely, the TC-CBF's exclusion regions are designed to facilitate smoother navigation around obstacles for nonholonomic vehicles, thereby enhancing efficiency.

\begin{remark}
The ED-CBF method introduces a second-order CBF that explicitly includes the velocity term, as formulated in \eqref{eq:edcbf}. In contrast, the TC-CBF approach uses a first-order CBF, which also incorporates velocity into its formulation. While these two approaches differ in order, both rely on the vehicle's velocity and position.
\end{remark}

\subsection{MPC with TC-CBF}
To generate a collision-free trajectory using the proposed TC-CBF, it is combined with MPC, resulting in the MPC-TCCBF framework. 
The integration of these methods utilizes the discrete-time version of the CBF introduced in Section~\ref{sec:subsec}, incorporating it as a constraint for obstacle avoidance.
The cost function for this strategy is formulated in the discrete-time domain as follows:
\begin{subequations}
\label{NMPC} %
\begin{equation}
\begin{aligned} \label{mpc:cost}
\min_{\mathbf{x}(\cdot),{\mathbf{u}}(\cdot)}\sum_{i=0}^{N-1} \Big(
|| \mathbf{x}_i - \mathbf{r}_i ||^2_Q  + || \mathbf{u}_i ||^2_R & + ||\Delta\mathbf{u}_i||^2_{R_d} \Big)  \\ &+ ||\mathbf{x}_{N} - \mathbf{r}_{N}||^2_{P} ,
\end{aligned}
\end{equation}
\begin{align}
    \text{s.t. } \mathbf{x}_0 - \mathbf{x}_{\text{init}} &= 0,  \label{mpc:init} \\
    \mathbf{x}_{i+1} - f_d(\mathbf{x}_i,\mathbf{u}_i) &= 0, \ i=0,\ldots,N-1, \label{mpc:dynamics} \\
    g(\mathbf{u}_i)  &\leq 0, \ i=0,\ldots,N-1, \label{mpc:constraints1} \\
     \Delta {h}_t(\mathbf{x}_i,\mathbf{u}_i) + \alpha_t h_t(\mathbf{x}_i) &\geq 0, \  i=0,\ldots,N-1,  \label{mpc:cbfo} 
\end{align}
\end{subequations}
where the matrices $Q$, $R$, $R_d$, and $P$ are weight matrices that penalize state error, control input, control input change rate, and terminal state error, respectively. 
The control input change rate is defined as $\Delta\mathbf{u}_i = (\mathbf{u}_i - \mathbf{u}_{i-1})/{T_s}$, where $T_s$ is the sampling time.
The reference state vector is denoted by $\mathbf{r}_i$, $N$ is the prediction horizon, $\mathbf{x}_{\text{init}}$ is the initial state, and $f_d$ represents the discretized dynamics function of the vehicle with the given sampling time. The constraints in \eqref{mpc:constraints1} define the inequality constraints for the control input and its change rate. 
The obstacle avoidance constraints for the TC-CBF are specified in \eqref{mpc:cbfo}. 
The difference in CBF is expressed as
$\Delta {h}_t(\mathbf{x}_i,\mathbf{u}_i) = {h}_t(\mathbf{x}_{i+1}) - {h}_t(\mathbf{x}_i)$, implying that ${h}_t(\mathbf{x}_{i+1}) \geq (1-\alpha_t){h}_t(\mathbf{x}_{i})$. 
This formulation ensures that the lower bound of the proposed CBF decreases exponentially at a rate of $(1-\alpha_t)$.

The solution to \eqref{NMPC} yields a sequence of control inputs, of which only the first input is implemented at each time step. This constrained finite-time optimal control problem is solved iteratively at each subsequent time step based on the newly estimated state. The nonlinear programming problem formulated in \eqref{NMPC} is solved using CasADi \cite{casadi} and acados \cite{verschueren2022acados}.

\section{Experimental Validation} \label{section4}
To evaluate the effectiveness of the MPC-TCCBF approach for nonholonomic vehicles, we conducted a comparative analysis with the MPC-EDCBF approach. In the MPC-EDCBF method, the constraints in \eqref{eq:cbfcon} are used for obstacle avoidance instead of those in \eqref{mpc:cbfo}.

\subsection{Unicycle simulations}
\subsubsection{Simulation setting}
Simulations were conducted using the nonholonomic unicycle model defined in \eqref{unicycle}.
The control frequency was set to \SI{10}{Hz}, with a sampling time of \SI{0.1}{\second}. The prediction horizon was set to $N=10$. 
The weight matrices used in the MPC were specified as $Q = \text{diag}([0,\ 2,\ 25, \ 100])$, $R = \text{diag}([50,\ 50])$, $P = \text{diag}([0,\ 2,\ 25, \ 100])$, and $R_d = \text{diag}([ 5,\ 5])$. These values were selected through trial and error to optimize both obstacle avoidance and path-tracking performance. 
The desired velocity for the vehicle was set to $u_r = \SI{2.0}{\meter/\second}$.
The initial state and control input were defined as $\mathbf{x}_{\text{init}} = [0,\ 0,\ 0,\ u_r]^\top$ and $\mathbf{u}_{\text{init}} = [0,\ 0]^\top$.
Three scenarios were simulated: static obstacle, head-on obstacle, and overtaking scenario. 
In all scenarios, the starting position was set at $(0,\ 0)$, with the target destination positioned at $(40,\ 0)$, $(50,\ 0)$, and $(40,\ 0)$, respectively. In the static obstacle scenario, an obstacle was placed at $(15,\ 0)$ with a radius of $o_r = \SI{2}{\meter}$. The head-on obstacle scenario involved an obstacle starting at $(30,\ 0)$ with a radius of $o_r = \SI{1}{\meter}$, moving at $\SI{-0.75}{\meter/\second}$. In the overtaking scenario, an obstacle started at $(10,\ 0)$ with a radius of $o_r = \SI{1}{\meter}$, moving at $\SI{0.5}{\meter/\second}$.

\subsubsection{Parameter Settings}
Initially, we adjusted two parameters of the ED-CBF in a static obstacle avoidance scenario. Specifically, we varied $\alpha$ from $0.25$ to $1.00$, and $\alpha_e$ from $0.03$ to $0.07$.
As shown in Fig.~\ref{fig:static_param_test}, 
all parameter combinations led to both aggressive avoidance maneuvers and speed reductions. 
Larger values of $\alpha$ lead to more aggressive avoidance maneuvers and significant speed reductions, while smaller values result in more conservative trajectories. The parameter $\alpha_e$ controls the steepness of the gradient constraints of the CBF: smaller values led to conservative maneuvers, whereas larger values induced more aggressive behavior, similar to the effect of $\alpha$. Among the parameter sets tested, $\alpha=0.5$ and $\alpha_e=0.05$ provided the best performance and were thus used in all subsequent comparison scenarios. The TC-CBF parameter $\alpha_t$ was set to $0.05$, matching $\alpha_e$.

\subsubsection{Results}
\begin{figure}[t]
\captionsetup[subfigure]{justification=centering}
    \centering
    \begin{subfigure}[h]{\linewidth}
        \centering
        \includegraphics[width=\textwidth]{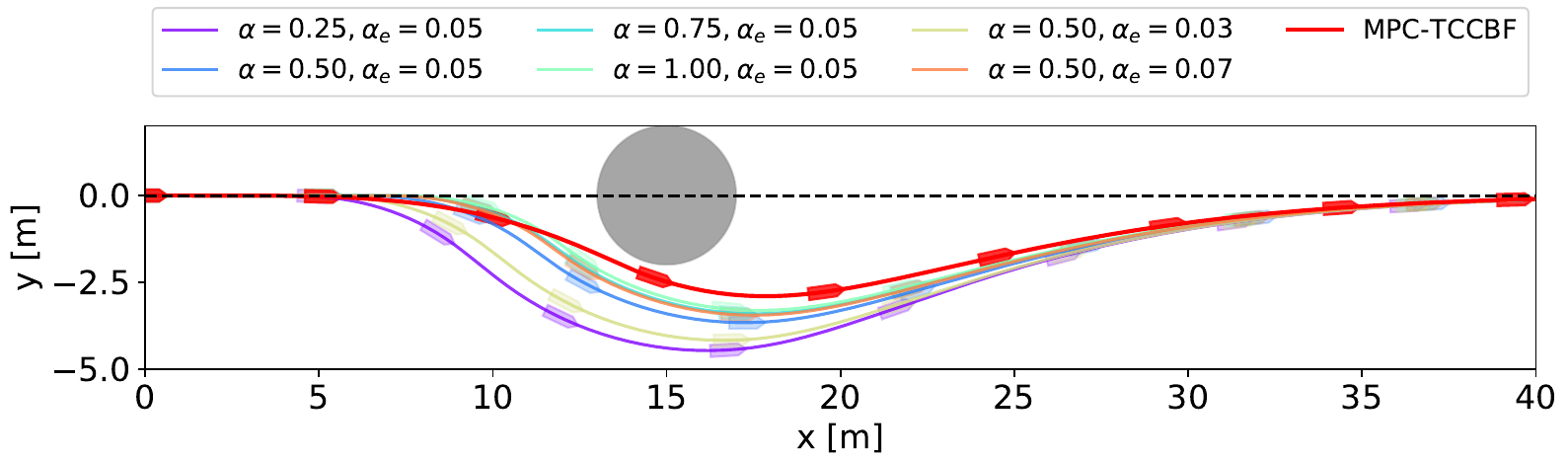}
        \caption{Trajectories.}
    \end{subfigure}
    \begin{subfigure}[h]{\linewidth}
        \centering
        \includegraphics[width=\textwidth]{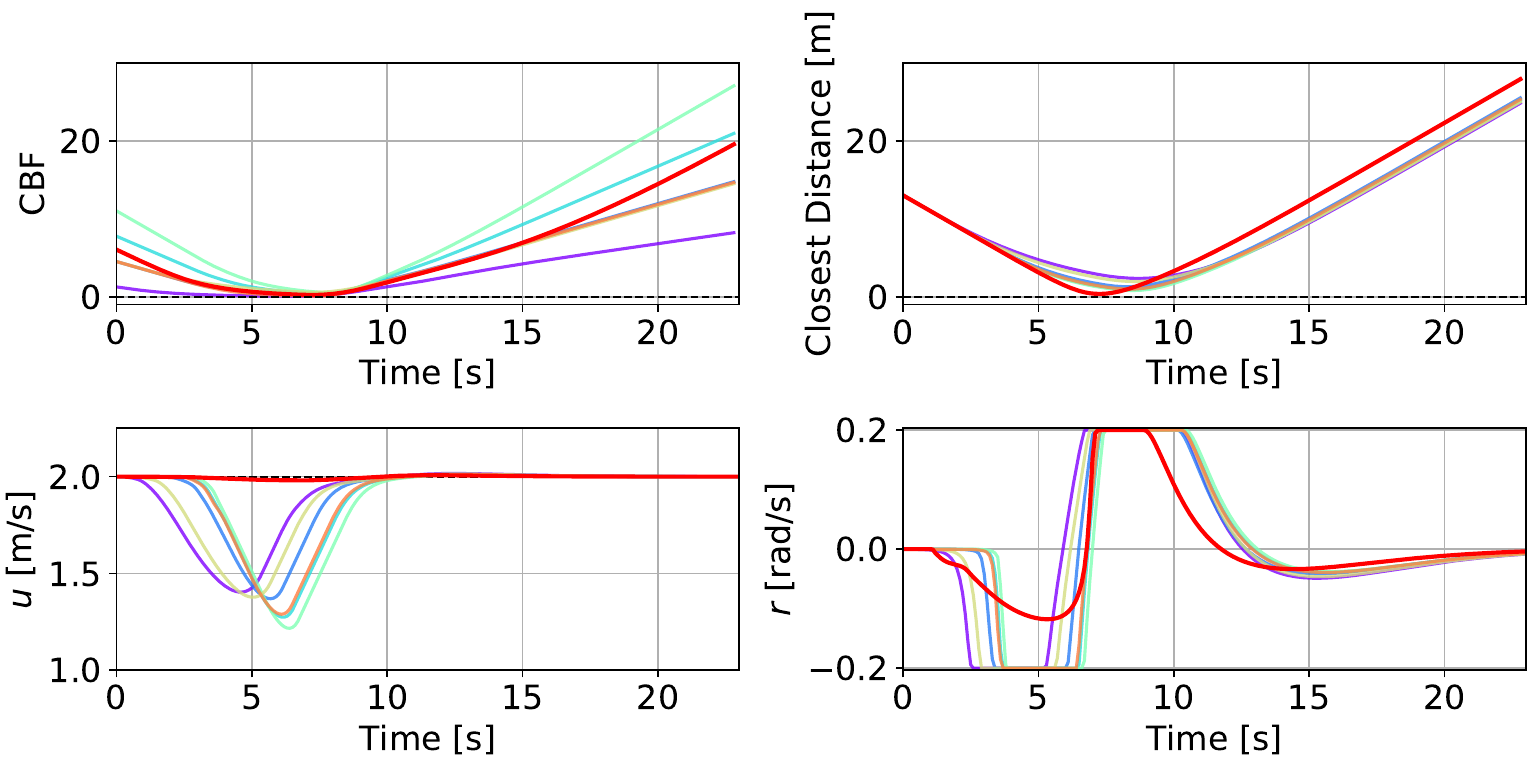}
        \caption{Time trajectories of CBF values, closest distances to the obstacle, velocities, and control inputs.}
    \end{subfigure}
    \caption{Result comparison with different parameter sets in the static obstacle avoidance scenario.}
    \label{fig:static_param_test}
\end{figure}

\begin{figure}[t]
\captionsetup[subfigure]{justification=centering}
    \centering
    \begin{subfigure}[h]{\linewidth}
        \centering
        \includegraphics[width=\textwidth]{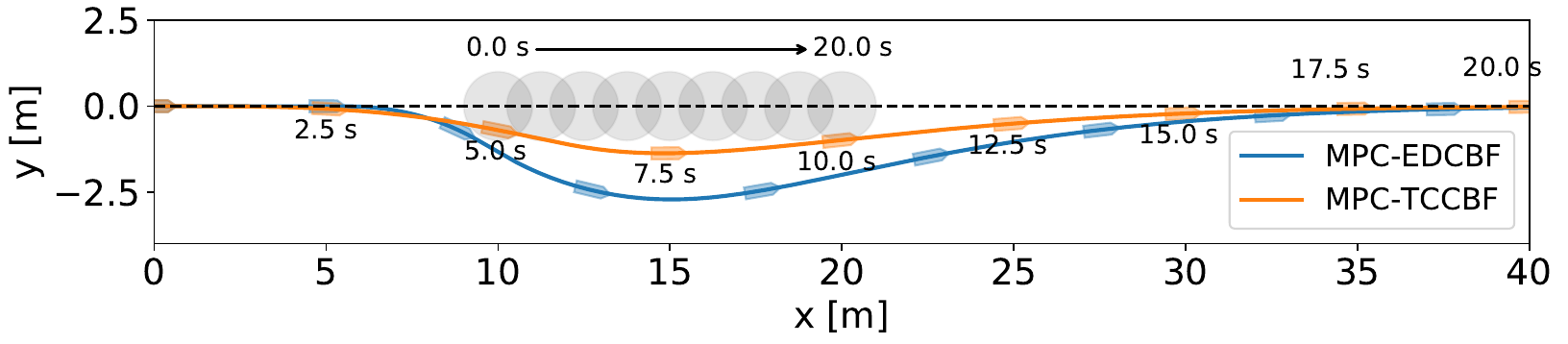}
        \caption{Trajectories.}
    \end{subfigure}
    \begin{subfigure}[h]{\linewidth}
        \centering
        \includegraphics[width=\textwidth]{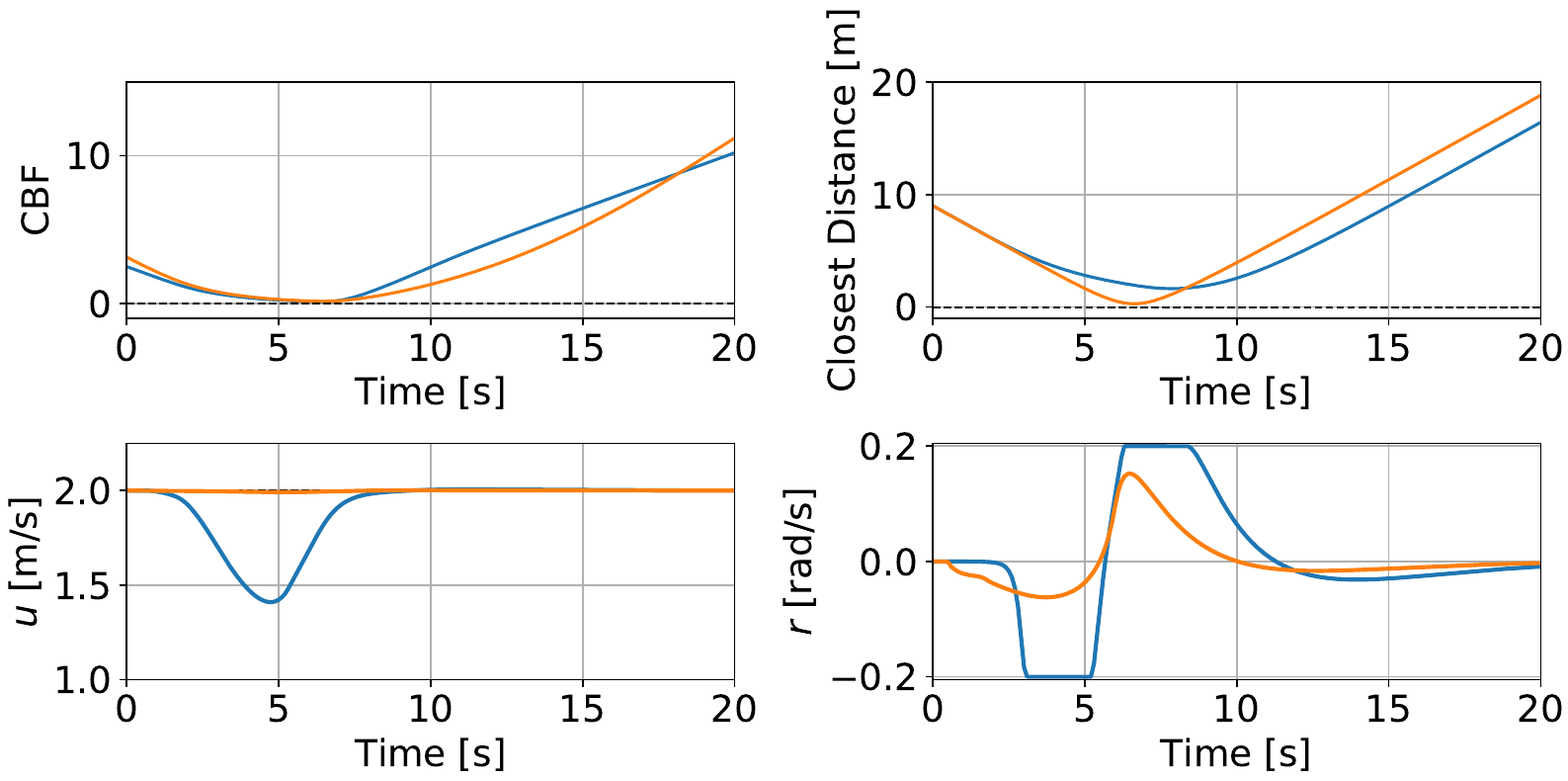}
        \caption{Time trajectories of CBF values, closest distances to the obstacle, velocities, and control inputs.}
    \end{subfigure}
    \caption{Result comparison in the overtaking scenario.}
    \label{fig:overtaking}
\end{figure}

Figures~\ref{fig:static_param_test} and \ref{fig:overtaking} show the results for the static obstacle avoidance and overtaking scenarios. These figures illustrate the vehicle's trajectory throughout the simulations, with snapshots captured every \SI{2.5}{\second}, highlighting changes in CBF values, the closest distance to the obstacle, velocities, and rotational speeds.
In both scenarios, the MPC-EDCBF algorithm induces unnecessary deceleration and aggressive evasive maneuvers during obstacle avoidance.
This deceleration occurs because the ED-CBF constraint cannot be satisfied while maintaining the desired velocity, resulting in a speed reduction when approaching the obstacle.
In contrast, the TC-CBF approach enables efficient obstacle avoidance by allowing smaller, smoother control inputs while maintaining desired speeds across all scenarios.
For a comprehensive comparative analysis, metrics such as arrival time $t_a$, average speed tracking error $e_{\text{speed}}$, and average cross-track error $e_{\text{cte}}$ were considered. Arrival time is defined as the moment the vehicle reaches the target $x$-position. The results, summarized in Table~\ref{table:unicycle}, demonstrate that the proposed approach consistently outperforms the ED-CBF across all metrics.

\begin{table}[t]
\centering
\small
\renewcommand{\arraystretch}{1.43}
    \begin{tabular}{ccccccc}
    \hline
    Scenario & 
    Controller & 
    $t_a$ & 
    $e_{\text{speed}}$ & 
    $e_{\text{cte}}$
    \\ \hline 
    \multirow{2}{*}{\makecell{Static}}  
    & MPC-EDCBF
    & 21.6
    & 0.088
    & 1.273
    \\ \cdashline{2-7}  
    & MPC-TCCBF
    & \bf{20.4}
    & \bf{0.005}
    & \bf{0.962}
    \\ \hline 
    \multirow{2}{*}{\makecell{Head-on}}  
    & MPC-EDCBF
    & 26.9
    & 0.107
    & 0.889
    \\ \cdashline{2-7}  
    & MPC-TCCBF
    & \bf{25.5}
    & \bf{0.019}
    & \bf{0.659}
    \\ \hline 
    \multirow{2}{*}{\makecell{Overtaking}}  
    & MPC-EDCBF
    & 21.3
    & 0.087
    & 0.916
    \\ \cdashline{2-7}  
    & MPC-TCCBF
    & \bf{20.1}
    & \bf{0.002}
    & \bf{0.450}
    \\ \hline 
\end{tabular}
\caption{Comparative simulation analysis of the MPC-EDCBF and MPC-TCCBF approaches for a unicycle.}
\label{table:unicycle}
\end{table}

\subsection{Underactuated ASV experiments}
In this section, we extend the TC-CBF approach to underactuated ASVs. Unlike the unicycle model, ASVs introduce complexity due to their sideways speed component. Because of their underactuated nature, ASVs cannot fully stabilize sideways motion without sacrificing control authority in other directions. To overcome this challenge, we apply the TC-CBF approach to ASVs by considering both speed over ground (SOG) and course over ground (COG). 
Through real-world experiments, we demonstrate the effectiveness and generalizability of the proposed approach.

\subsubsection{Vehicle modeling}
The vehicle dynamics are defined in both body-fixed and inertial coordinate systems. The equations governing the dynamics are as follows \cite{fossen2011handbook}:
\begin{subequations}   \label{eq:dynamics}%
   \begin{gather}
    M\dot{\nu}+C(\nu )\nu +D(\nu)\nu = \tau_{c}, \label{eq:kinetics}\\ 
    \dot{\eta} = R(\psi)\nu , \\ 
    R(\psi) = \begin{bmatrix}
    \cos \psi  & -\sin\psi & 0 \\
    \sin\psi & \cos\psi & 0 \\
    0 & 0 &  1
    \end{bmatrix},
    \label{eq:rot}
   \end{gather}
\end{subequations}
where $ \nu = [u,\,v,\,r]^\top$ and $\eta = [x,\,y,\,\psi]^\top$ represent the velocity and position vectors, respectively. The state vector is defined as $\mathbf{x} = [\eta^\top, \nu^\top]^\top$. Here, $M$ denotes the inertia matrix, $C(\nu)$ is the Coriolis-centripetal matrix, $D(\nu)$ is the damping matrix, and $\tau_c = [\tau_X,0,\tau_N]^\top$ represents the control forces and moment. The rotation matrix $R(\psi)$ transforms coordinates from the body-fixed frame to the inertial frame.
Assuming symmetry of the vehicle along the $x$ and $y$ directions, the matrices $M$, $C(\nu)$, and $D(\nu)$ are defined as follows:
\begin{subequations}
\begin{align}
M &= \text{diag}(m_{11},m_{22},m_{33}),
    \label{eq:M_matrix}  \\
C(\nu) &= \begin{bmatrix} 
        0 & 0 & -m_{22}v \\
        0 & 0 &  m_{11}u \\
        m_{22}v & -m_{11}u & 0
\end{bmatrix},    \label{eq:C_matrix}\\
D(\nu) &= -\begin{bmatrix} 
        X_u+X_{u|u|}|u| & 0 & 0 \\
        0 & Y_v + Y_{v|v|}|v|& Y_r \\
        0 & N_v & N_r + N_{rrr}r^2
\end{bmatrix}. \nonumber \\
\label{eq:D_matrix}
\end{align}
\label{eq:MCD_matrix}
\end{subequations}

\subsubsection{TC-CBF for ASVs}
The vehicle's kinematic model is redefined using the SOG $V$ and COG $\psi_w$:
\begin{equation}
    \dot{x} =  V \cos \psi_w, \
    \dot{y} =  V \sin \psi_w, \
    \dot{\psi}_w = r + \dot{\beta},
\end{equation}
where $\psi_w = \psi + \beta$, $\beta = \arctan (v/u)$, and $V = \sqrt{u^2+v^2}$. The radius of the turning circle is calculated as:
\begin{equation}
    R = \frac{V}{r_{\text{max}}},
\end{equation}
where $r_{\text{max}} = \max (\dot{\psi}_w)$.
The two turning circle centers defined in \eqref{eq:tccbf_form} are redefined based on the COG as follows:
\begin{equation}
\begin{aligned}    
    (x_{tr},y_{tr})  &= (x + R\cos(\psi_w-\pi/2), y+R\sin(\psi_w-\pi/2)), \\
    (x_{tl},y_{tl})  &= (x + R\cos(\psi_w+\pi/2), y+R\sin(\psi_w+\pi/2)).
\end{aligned}
\end{equation}

\begin{figure}[t]       
    \centerline{\includegraphics[width=0.64\linewidth]{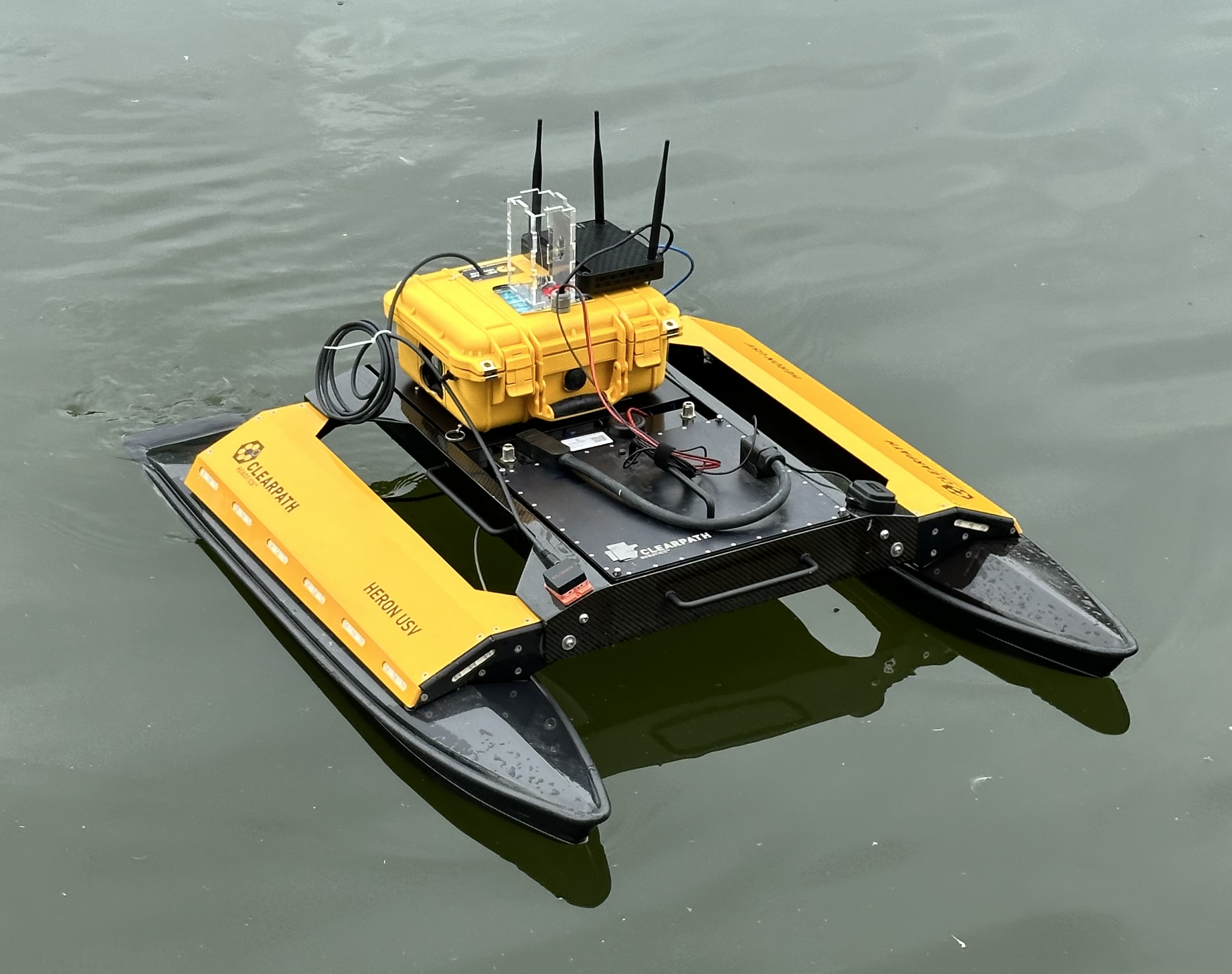}}
    \caption[heron.]{The Clearpath Robotics Heron ASV used in the experiments.}     
    \label{fig:heron}
\end{figure}

\begin{figure}[t!]
\captionsetup[subfigure]{justification=centering}
    \centering
    \begin{subfigure}[h]{0.95\linewidth}
        \centering
        \includegraphics[width=\textwidth]{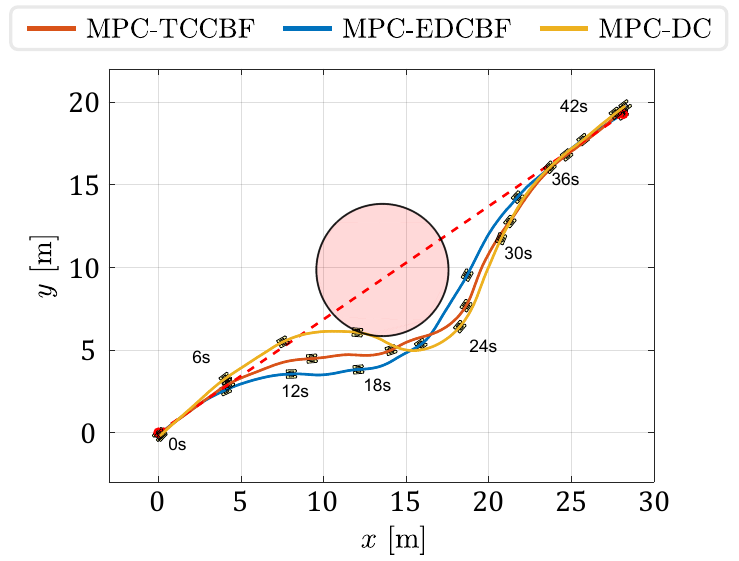}
        \caption{Trajectories.}
        \label{fig:heron_static_traj}
    \vspace{0.25cm}
    \end{subfigure}    
    \begin{subfigure}[h]{0.81\linewidth}
        \centering
        \includegraphics[width=\textwidth]{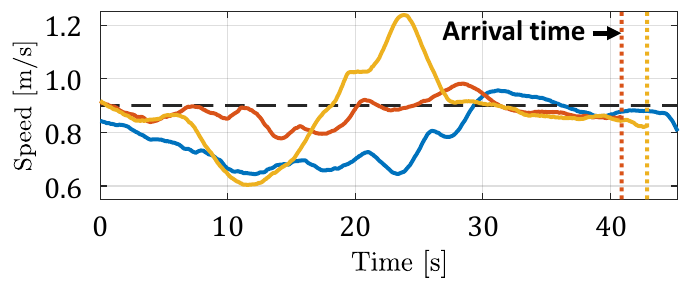}
        \caption{Time trajectories of velocities.}
        \label{fig:heron_static_vel}
    \end{subfigure}
    \begin{subfigure}[h]{0.8\linewidth}
        \centering
        \includegraphics[width=\textwidth]{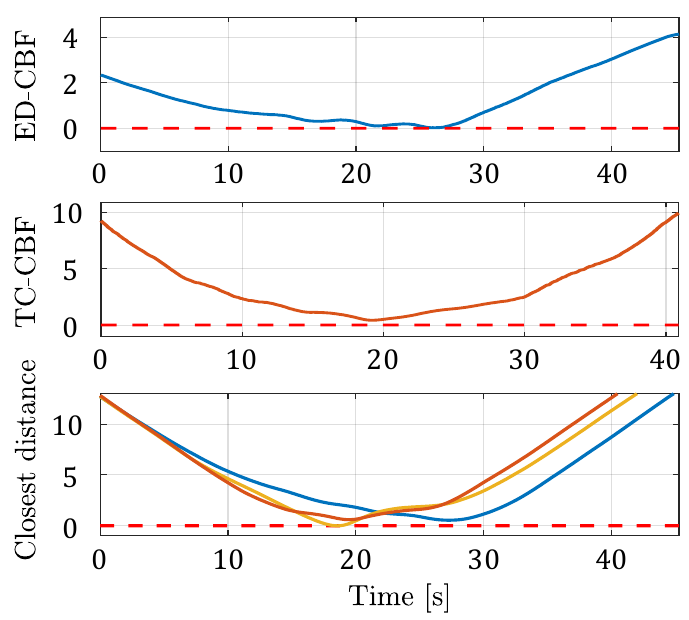}
        \caption{Time trajectories of CBF values and closest distances to the obstacle.}
        \label{fig:heron_static_cbf_cd}
    \end{subfigure}
    \caption{Result comparison in the static obstacle avoidance scenario.}
    \label{fig:heron_static}
\end{figure}

\begin{figure*}[t!]
\captionsetup[subfigure]{justification=centering}
\centering
    \begin{subfigure}[h]{0.66\linewidth}
        \centering
        \includegraphics[width=\textwidth]{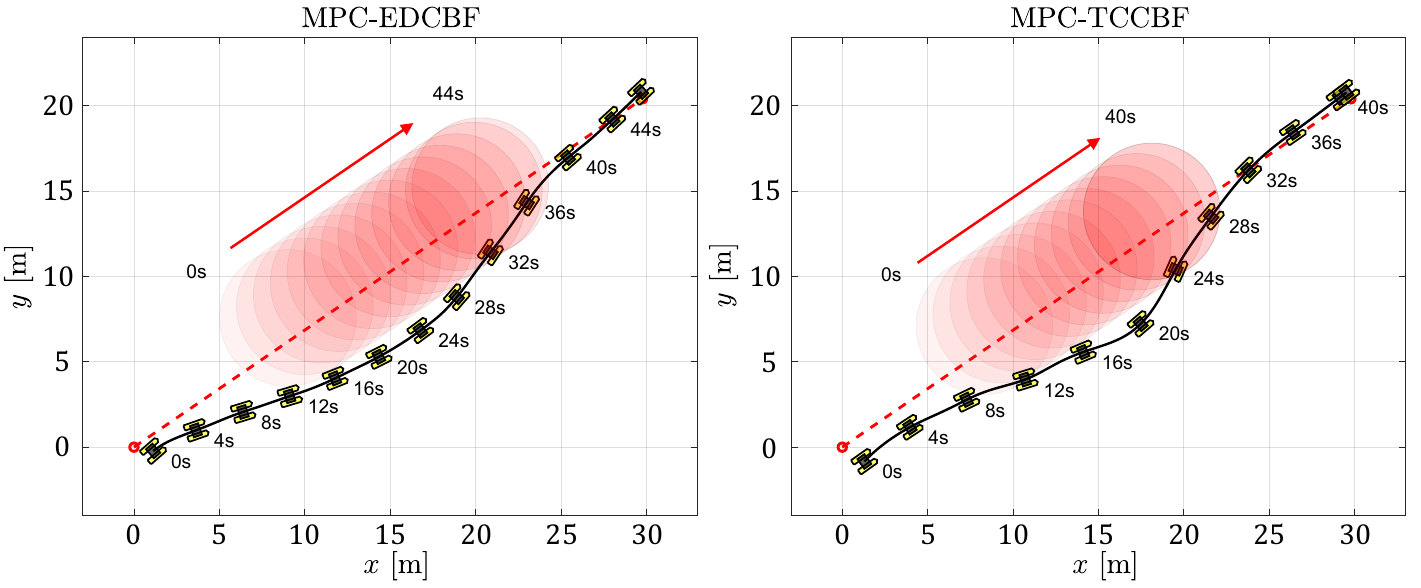}
        \caption{ASV and obstacle trajectories.}
        \label{fig:heron_overtaking_traj}
    \vspace{0.3cm}
    \end{subfigure}
    \begin{subfigure}[h]{0.28\linewidth}
        \centering
        \includegraphics[width=\textwidth]{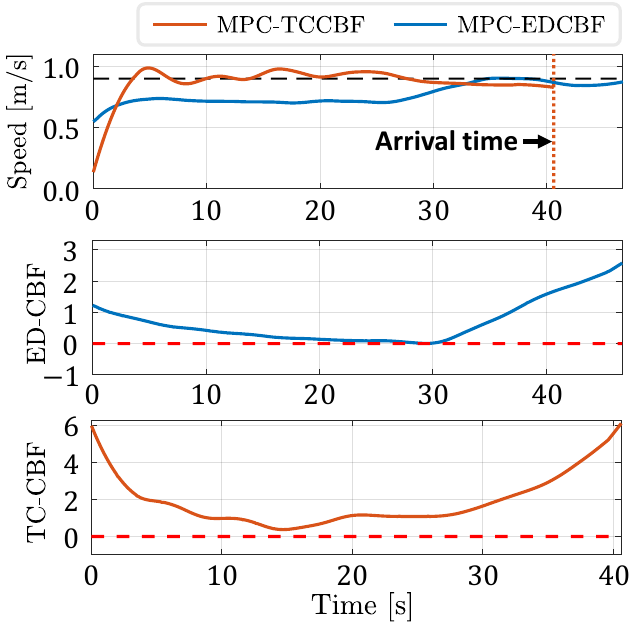}
        \caption{Velocities and CBF values.}
        \label{fig:heron_overtaking_cbf_vel}
    \vspace{0.1cm}
    \end{subfigure}   
    \vspace{0.15cm}
    \begin{subfigure}[h]{0.49\linewidth}
        \centering
        \includegraphics[width=\textwidth]{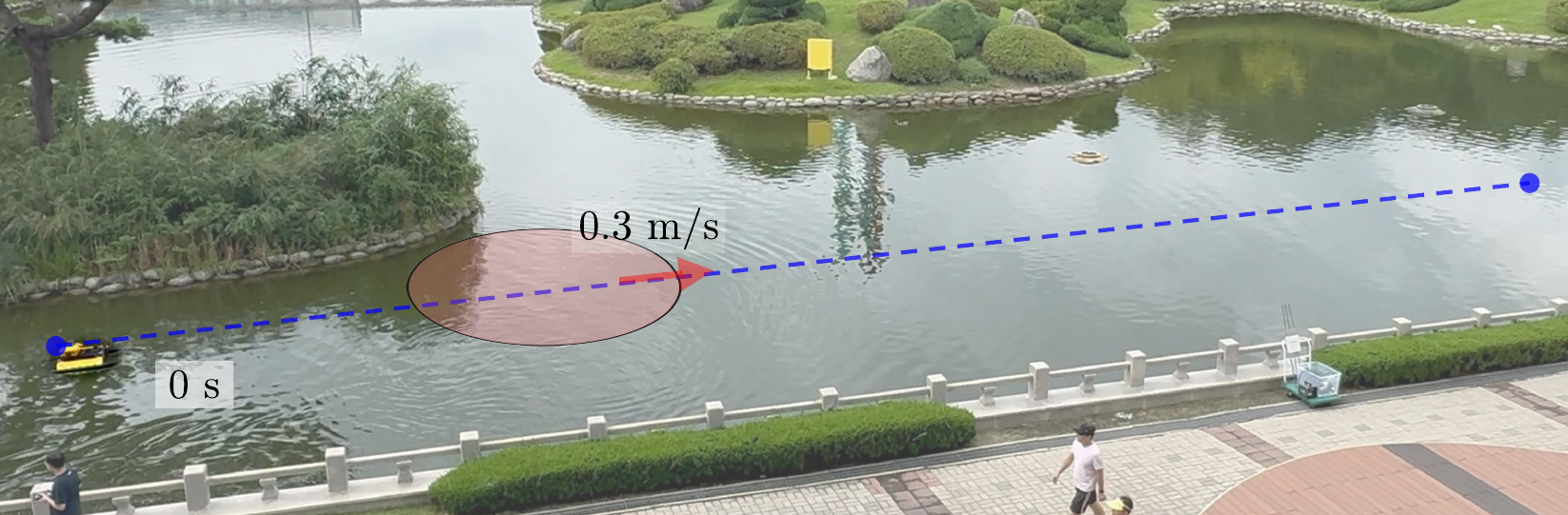}
        \caption{Results at \SI{0}{\second}.}
        \label{fig:heron_overtaking_0s}
    \end{subfigure}  
    \begin{subfigure}[h]{0.49\linewidth}
        \centering
        \includegraphics[width=\textwidth]{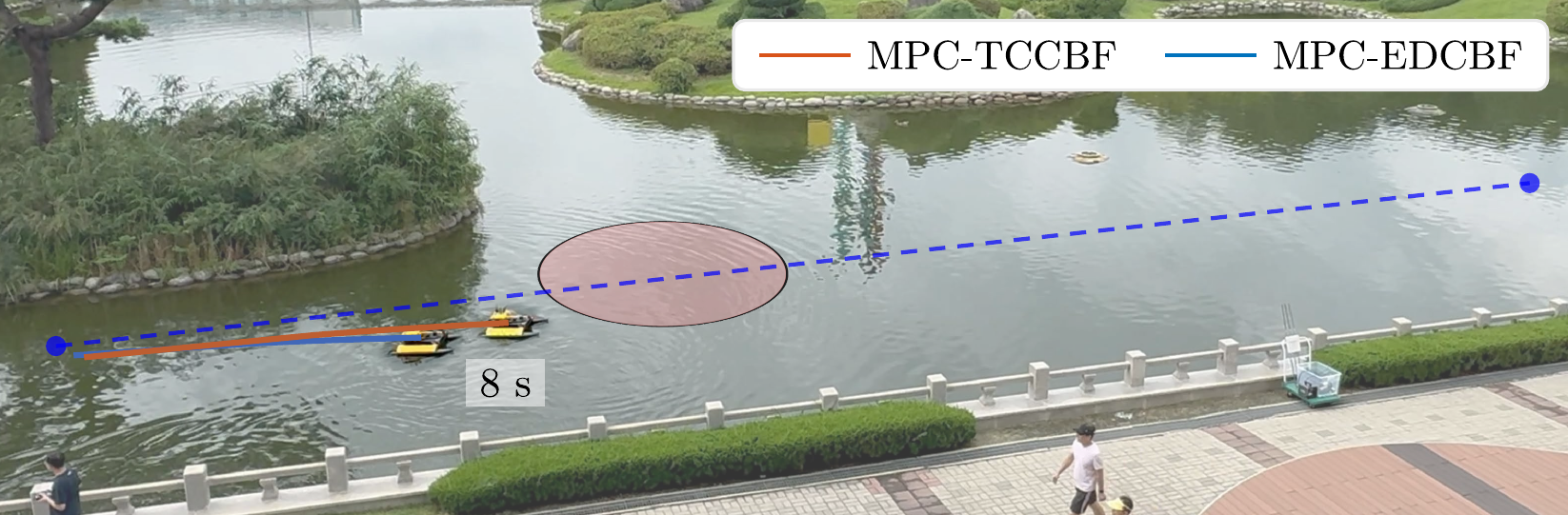}
        \caption{Results at \SI{8}{\second}.}
    \end{subfigure}   
    \vspace{0.15cm}
    \begin{subfigure}[h]{0.49\linewidth}
        \centering
        \includegraphics[width=\textwidth]{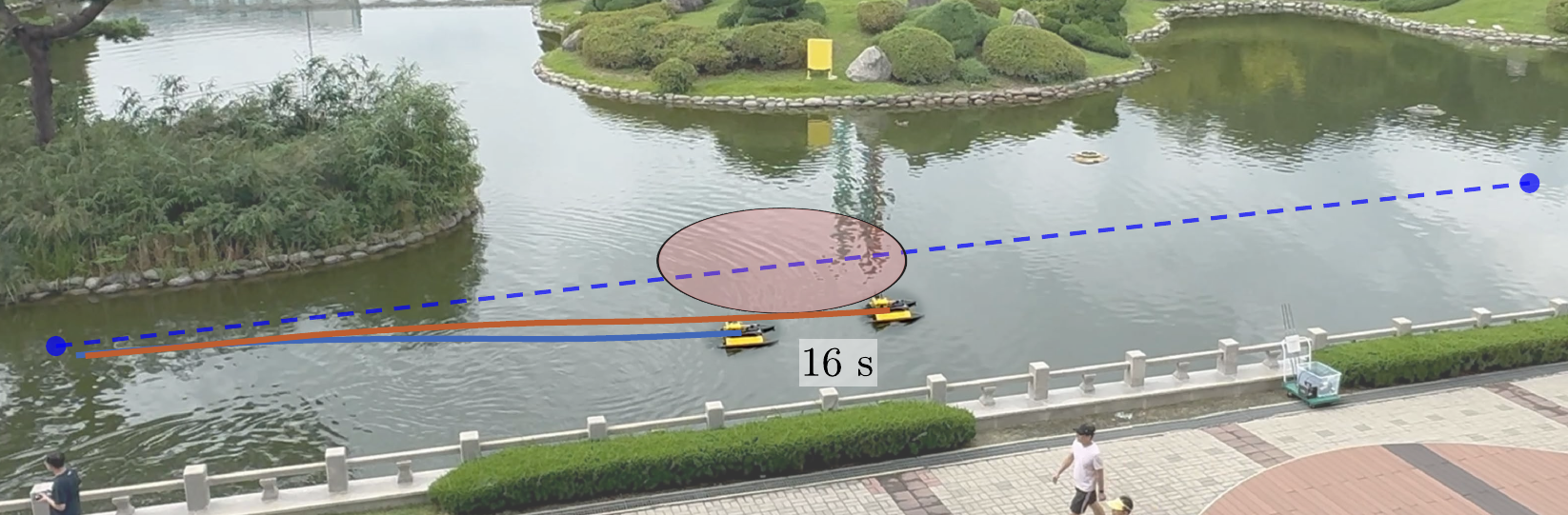}
        \caption{Results at \SI{16}{\second}.}
    \end{subfigure}     
    \begin{subfigure}[h]{0.49\linewidth}
        \centering
        \includegraphics[width=\textwidth]{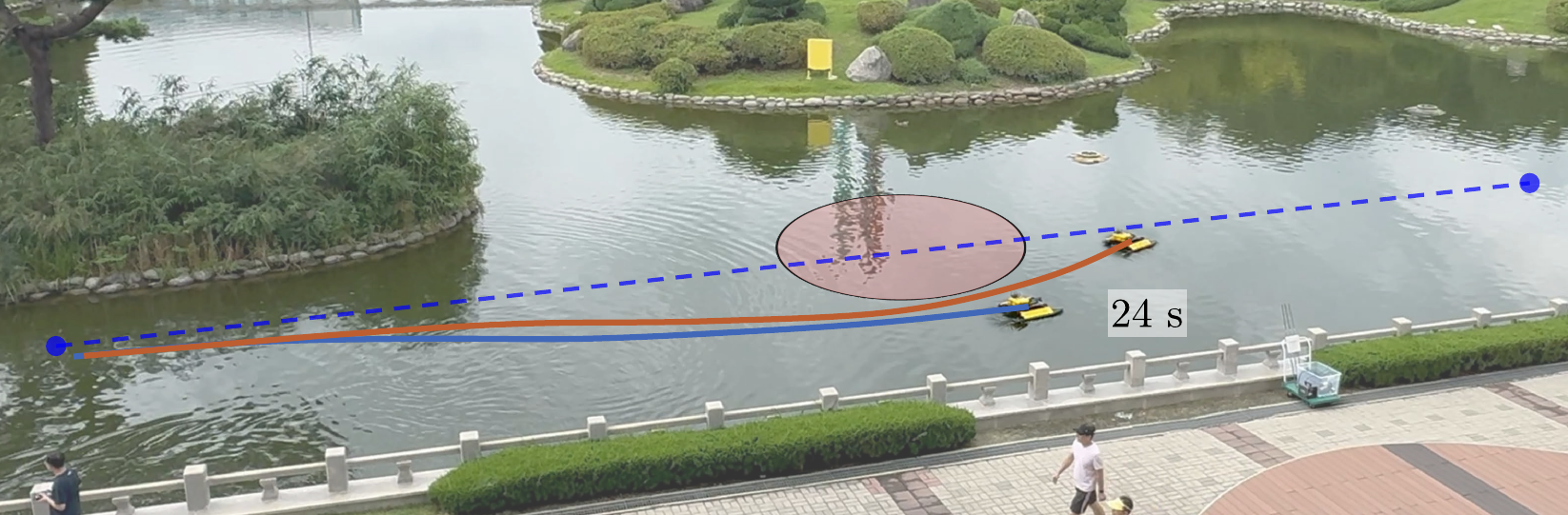}
        \caption{Results at \SI{24}{\second}.}
    \end{subfigure}   
    \begin{subfigure}[h]{0.49\linewidth}
        \centering
        \includegraphics[width=\textwidth]{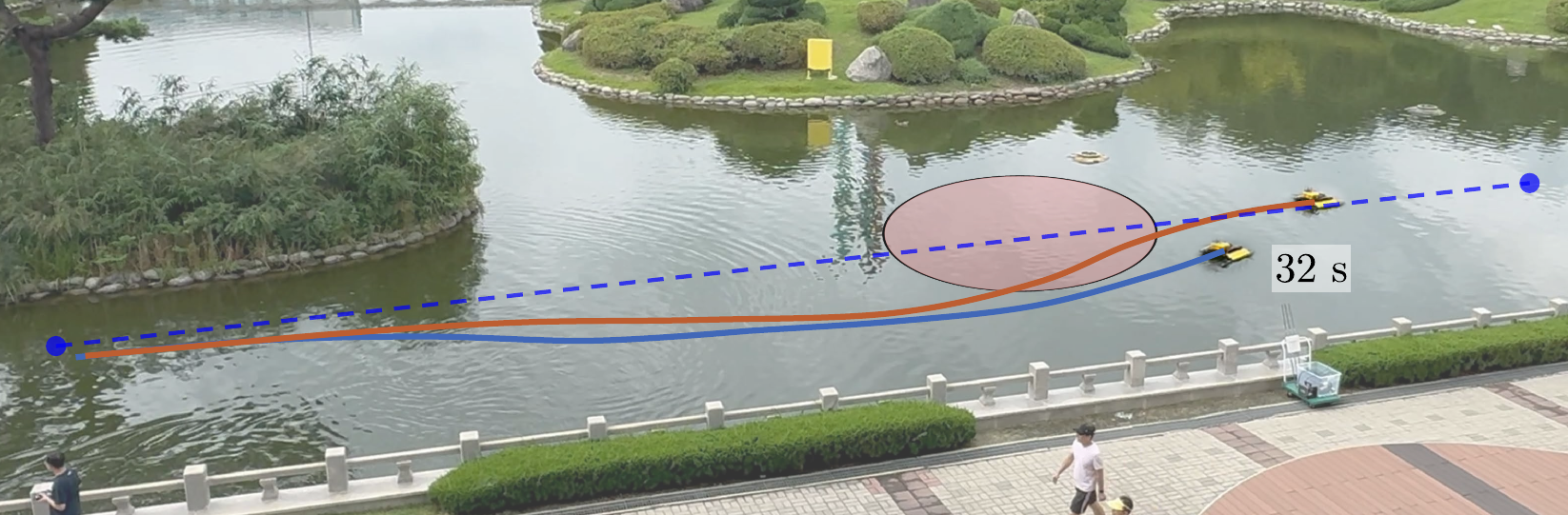}
        \caption{Results at \SI{32}{\second}.}
    \end{subfigure}         
    \begin{subfigure}[h]{0.49\linewidth}
        \centering
        \includegraphics[width=\textwidth]{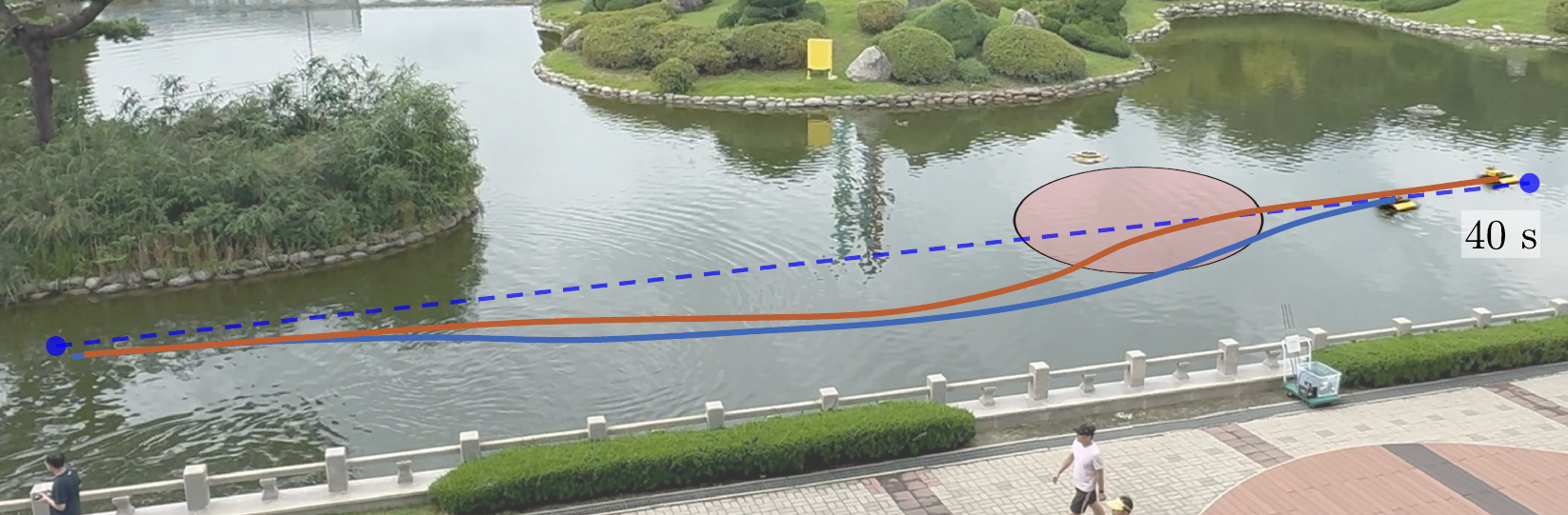}
        \caption{Results at \SI{40}{\second}.}
        \label{fig:heron_overtaking_40s}
    \end{subfigure}         
    \caption{Results of the experimental run in the overtaking scenario using the two algorithms, with obstacle areas highlighted in red. Snapshot images were captured at 8-second intervals.}
    \label{fig:heron_overtaking}  
\end{figure*}


\subsubsection{Experimental settings}
To evaluate the performance of our approach, we used the Clearpath Robotics Heron ASV, as shown in Fig.~\ref{fig:heron}. The Heron is a 1.35 m × 0.98 m × 0.32 m catamaran-style vehicle equipped with parallel differential thrusters. The control forces and moments are defined as:
\begin{equation}        
    \tau_X = F_l + F_r, \
    \tau_N = (-F_l+F_r)l,
\end{equation} 
where $l = \SI{0.3}{\meter}$ is the distance from the vehicle's center of gravity to the point of thrust application. The control input vector is defined as $\mathbf{u} = [F_l, F_r]^\top$, where $F_l$ and $F_r$ represent the forces generated by the left and right thrusters, respectively.

The ASV is equipped with an IMU, GPS, and a WiFi antenna, and features an onboard computer for managing navigation filters and optimization problems. 
Nonlinear optimization-based system identification was used to estimate the hydrodynamic parameters in \eqref{eq:MCD_matrix} \cite{lee2023nonlinear}. 

The control frequency was set to \SI{10}{Hz}, and the prediction horizon was established at $N=20$ with a sampling time of \SI{0.1}{\second}, resulting in a prediction time of \SI{2}{\second}.
The weight matrices used in the MPC were specified as $Q = \text{diag}([0,\ 1,\ 3, \ 50, \ 0, \ 3])$, $R = \text{diag}([10^{-6},\ 10^{-6}])$, $P = \text{diag}([0,\ 5,\ 15, \ 250, \ 0, \ 15])$, and $R_d = \text{diag}([0.03,\ 0.03])$. These values were selected through trial and error to optimize both obstacle avoidance and path-tracking performance.
The desired velocity for the vehicle was set at $u_r = \SI{0.9}{\meter/\second}$.
Similar to the simulation studies, three scenarios were considered in the experiments. The obstacle radius was set at $\SI{4}{\meter}$, moving at $\SI{0.3}{\meter/\second}$ in dynamic scenarios. 

The CBF parameters were set to $r_{\text{max}} = 0.2$ and $\alpha = 1$ to ensure that the depth of the restricted area when facing the obstacle head-on is similar across both CBFs, as illustrated in Fig.~\ref{fig:cbfcompare}. Tests were conducted by varying $\alpha_e$ and $\alpha_t$ to 0.01, 0.015, and 0.02. For $\alpha_t$, a feasible avoidance trajectory was found in all three scenarios across all parameter settings, with 0.02 demonstrating the best performance, so it was selected. The smoothing parameter was set to $k=5$. For $\alpha_e$, a value of 0.02 resulted in an infeasible state in head-on static scenarios; therefore, 0.015, which performed better than 0.01, was chosen for the experiments.

\subsubsection{Results}
Figure~\ref{fig:heron_static} presents the results from the static obstacle avoidance scenario. Figures~\ref{fig:heron_static_traj}-\ref{fig:heron_static_cbf_cd} show various aspects of the vehicle's behavior over time, including the trajectory with snapshots taken every \SI{6}{\second}, time trajectories of velocities, CBF values, and the closest distance to the obstacle.
In this scenario, the distance constraint-based MPC algorithm (MPC-DC) was evaluated as well. 
The MPC-DC algorithm with $N=50$ and a sampling time of \SI{0.1}{\second} was unable to find feasible avoidance trajectories. In contrast, with $N=50$ and a sampling time of \SI{0.2}{\second}, the MPC-DC successfully identified feasible avoidance trajectories but initiated avoidance only when the vehicle was close to the obstacle boundary. This resulted in abrupt turns and speed reductions, despite the longer prediction horizon compared to the CBF-based algorithms.
On the other hand, both CBF-based approaches showed comparable results even with a shorter horizon. However, the MPC-EDCBF approach caused speed reductions, making it less efficient than the MPC-TCCBF approach. 

Figure~\ref{fig:heron_overtaking} presents the results of the overtaking scenario. The time trajectories of the ASVs and obstacles during the experiments are shown in Fig.~\ref{fig:heron_overtaking_traj}, with corresponding time-lapse views of the real-world experimental runs shown in Figs.~\ref{fig:heron_overtaking_0s}-\ref{fig:heron_overtaking_40s}.
As illustrated in Fig.~\ref{fig:heron_overtaking_cbf_vel}, the ED-CBF method exhibits unnecessary speed reductions during the overtaking process, impacting overall performance. In contrast, the TC-CBF approach proved more efficient by allowing faster overtaking while closely tracking the target speed. Quantitative analysis of the three scenarios, including the head-on scenario, was conducted using the same metrics as in the simulation study. The results, summarized in Table~\ref{table:ASV}, indicate that the proposed approach outperforms the ED-CBF method across all evaluated metrics. 
The proposed approach reduced mission completion time by up to 13\% while also enhancing path-following performance.
Further experimental results are available in the supplementary video (\url{https://youtu.be/_fUdCjV7Lbg}).

\begin{table}[h]
\centering
\small
\renewcommand{\arraystretch}{1.43}
    \begin{tabular}{ccccccc}
    \hline
    Scenario & 
    Controller & 
    $t_a$ & 
    $e_{\text{speed}}$ & 
    $e_{\text{cte}}$
    \\ \hline 
    \multirow{2}{*}{\makecell{Static}}  
    & MPC-EDCBF
    & 45.4
    & 0.124
    & 1.765
    \\ \cdashline{2-7}  
    & MPC-TCCBF
    & \bf{41.0}
    & \bf{0.040}
    & \bf{1.623}
    \\ \hline 
    \multirow{2}{*}{\makecell{Head-on}}  
    & MPC-EDCBF
    & 42.2
    & 0.097
    & 2.014
    \\ \cdashline{2-7}  
    & MPC-TCCBF
    & \bf{38.8}
    & \bf{0.051}
    & \bf{1.453}
    \\ \hline 
    \multirow{2}{*}{\makecell{Overtaking}}  
    & MPC-EDCBF
    & 46.8
    & 0.133
    & 2.113
    \\ \cdashline{2-7}  
    & MPC-TCCBF
    & \bf{40.7}
    & \bf{0.063}
    & \bf{1.776}
    \\ \hline 
\end{tabular}
\caption{Comparative experimental analysis of the MPC-EDCBF and MPC-TCCBF approaches for an ASV.}
\label{table:ASV}
\end{table}

\section{Conclusion} \label{section5}
This paper presents the TC-CBF approach to improve obstacle avoidance efficiency for nonholonomic vehicles by integrating MPC with TC-CBF. Unlike traditional methods that rely on Euclidean distance to assess proximity, the TC-CBF approach assesses proximity based on the positions of the vehicle's turning circles. This new method enables the generation of more efficient trajectories across various operational scenarios while ensuring safety. The effectiveness of the proposed methodology has been validated through both numerical simulations with unicycles and real-world experiments with ASVs in diverse scenarios.

\bibliographystyle{IEEEtran}
\bibliography{IEEEabrv,main}
\end{document}